\documentclass[conference]{IEEEtran}
\IEEEoverridecommandlockouts
\usepackage{graphicx}
\usepackage{float}
\usepackage{subfigure}
\usepackage{cite}
\usepackage{amsmath,amssymb,amsfonts}
\usepackage{algorithmic}
\usepackage{graphicx}
\usepackage{textcomp}
\usepackage{booktabs}
\usepackage{xcolor}
\usepackage{booktabs}
\usepackage{listings}
\usepackage{algorithm}
\usepackage{algorithmic}
\usepackage{setspace}
\usepackage{hyperref}

\def\BibTeX{{\rm B\kern-.05em{\sc i\kern-.025em b}\kern-.08em
    T\kern-.1667em\lower.7ex\hbox{E}\kern-.125emX}}
\begin{document}

\title{A Discrete-event-based Simulator for Distributed Deep Learning\\
}

\author{
Xiaoyan Liu$^{\dag}$, Zhiwei Xu$^{\ddag}$$^{\dag}$, Yana Qin$^{\dag}$, Jie Tian$^{\S}$\\
\small $^{\dag}$College of Data Science and Application, Inner Mongolia University of Technology, Hohhot, China.\\
\small $^{\ddag}$Institute of Computing Technology, Chinese Academy of Sciences, Beijing, China\\
\small $^{\S}$Department of Computer Science, New Jersey Institute of Technology, Newark, USA.\\
\small Email: \{liuxiaoyan\_2021,yana\_qin,\}@foxmail.com,xuzhiwei2001@ict.ac.cn,jt66@njit.edu
}


\maketitle

\begin{abstract}
New intelligence applications are driving increasing interest in deploying deep neural networks (DNN) in a distributed way. To set up distributed deep learning involves alterations of a great number of the parameter configurations of network/edge devices and DNN models, which are crucial to achieve best performances. Simulations measure scalability of intelligence applications in the early stage, as well as to determine the effects of different configurations, thus highly desired. However, work on simulating the distributed intelligence environment is still in its infancy. The existing simulation frameworks, such as NS-3, etc., cannot extended in a straightforward way to support simulations of distributed learning. In this paper, we propose a novel discrete event simulator, sim4DistrDL, which includes a deep learning module and a network simulation module to facilitate simulation of DNN-based distributed applications. Specifically, we give the design and implementation of the proposed learning simulator and present an illustrative use case.
\end{abstract}

\begin{IEEEkeywords}
Intelligence application; Simulator for distributed deep learning; NS-3-based discrete event simulator; Distributed data caching and learning 
\end{IEEEkeywords}

\section{Introduction}
With the breakthrough of Artificial Intelligence (AI), we are witnessing a booming increase in AI-based applications and services. The existing intelligent applications are computation intensive.  
Recently, to efficiently process a huge number of data in the edge network, the deployment of distributed deep learning models has rapidly developed, which aims to improve the speed and scalability of data calculations and reduce time-consuming tasks. However, configurations of edge/network devices and deep learning models are quite challenging, which have brought great difficulties to the verification of new intelligent applications. Among three methods in response to this problem, theoretical analysis, simulation, and experimental deployment, simulation is an intermediate stage between the theoretical analysis and experimental deployment, and thus more flexible and efficient. It can be used as preliminary verification to support qualified implementation of the new applications of distributed deep learning \cite{r2}\cite{r3}. 
\begin{figure}[h]
\centerline{\includegraphics[width=\linewidth ]{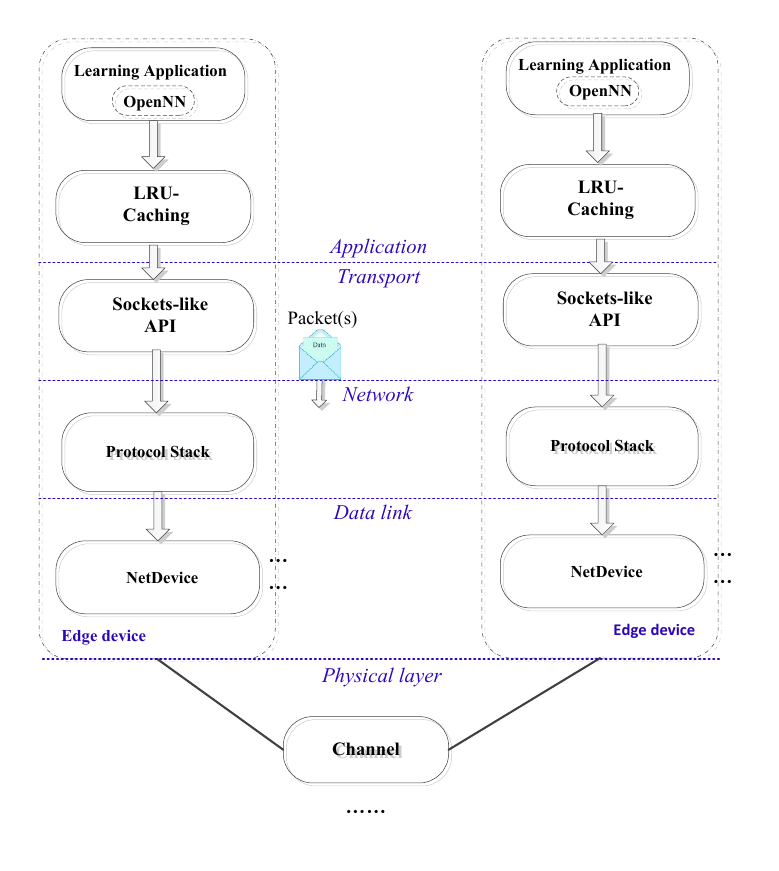}}
\vspace{-8pt}
\caption{Architecture of sim4DistrDL }
\label{fig0}
\end{figure}

Lack of the consideration of the computation logic inside of applications, the existing simulators used in the distributed computation environment, NS-3, MatLab, etc., are difficult to evaluate distributed deep learning. 
Among them, MatLab does not support systematic simulation methods considering discrete events\cite{r4}\cite{r5}, so that the effectiveness of distributed deep learning simulation on MatLab cannot be guaranteed. Similarly, NS-3 is lack of the support for the embed computational components. The deep learning libraries (Pytorch, Tensorflow, etc.) can not be embed in NS-3 in a straigtforward way. Hence, a discrete simulator for distributed deep learning is highly desired. 

To solve the existing problems and facilitate large-scale distributed training on the simulation platform, we design a discrete-event-based simulator for distributed deep learning, namely,  sim4DistrDL$\footnote{Online repository: \href{https://github.com/Lxy85/sim4DistrDL.git}{https://github.com/Lxy85/sim4DistrDL.git}}$. It provides a fundamental discrete event simulation framework, as well as a deep learning lib to enable various deep learning models, \emph{e.g.}, MLP, CNN, etc. (see Fig. 1). The proposed simulator will be very useful utilities to allow users to estimate times
and the costs for model training and inference before deploying the model. 

The main contributions of this study are summarized as follows:
\begin{itemize}
\item We leverage NS-3 to develop a discrete-event network simulator, sim4DistrDL. The essential networking components for distributed deep learning, such as LRU-based caching component, are designed and embed in NS-3. 
\item We achieve cross compiling of networking components and a deep learning library (OPENN) in the simulator, enable the support for the popular deep learning designs, such as MLP, CNN, etc.. Its applications in classification, regression, prediction and other artificial intelligence come to be feasible.
\item To evaluate whether sim4DistrDL can support the training and inference of DNN models, we provide application examples of sim4distributed DL. The evaluation results demonstrate that sim4distributed DL facilitates users to build learning environments for distributed trainning and simulate various DNN-based intelligence applications in a distributed way.
\end{itemize}

This paper is organized as follows. We study related work in  Section 2. The architecture of sim4distributed DL is present in Section 3, as well as detailed implementation of sim4distributed DL in Section 4. Application cases are provided to illustrate the use of sim4distributed DL. Finally, we conclude this work in Section 6.

\section{RELATED WORK}

As a brand-new technology under development, there are many well-known network simulators that provide a simulation environment, such as OPNET, QualNet, and REAL. However, they cannot be widely used due to restrictions in terms of use or high R\&D costs.

Previously, the performance verification of distributed deep learning schemes is still based on Matlab simulation. Initially, Yuqing Du et al.\cite{r6} conducted research and verification on the new hierarchical gradient quantization framework on the MatLab simulation platform. Later, Wen Sun et al.\cite{r7} integrated intelligence into distributed  computing and verified the proposed IIoT intelligent computing architecture on the MatLab platform. Ali Hassan Sodhro et al.\cite{r8} also used MatLab to emulate deep learning. However, MatLab lacks systematic simulation methods, such as discrete event simulation, cannot accurately characterize the impact of network transmission performance on the learning process in a distributed environment, and cannot guarantee the validity of distributed deep learning performance verification.


NS-3~\cite{r4} is a open-source, cross-platform discrete-event network simulator for Internet systems\cite{r11}. It supports distributed simulation and parallel simulation and runs faster. NS-3 is designed as a set of libraries that can be combined with other external software libraries\cite{r12}\cite{r16}. It became a standard in networking research as the results obtained were accepted by the community. However, NS-3 is lack of the support for the embed computational components. Although new discrete simulators based on deep reinforcement learning (DRL) or deep reinforcement learning (DL) have been proposed to satisfy the aforementioned requirements, recently. There are also some problems in these frameworks. Piotr Gawłowicz et.al \cite{r13} have developed a module, NS3-gym, to provide an interface in Reinforced Learning between NS-3 and OpenAI Gym. They use ZMQ socket to implement the communication between NS-3 and OpenAI Gym. However, facing the next generation of networks, the network's density is rapidly increasing. A large amount of data may require to transmit from NS-3 to AI modules for training or testing. It is crucial to reduce the time-consuming and support colossal data transmission, which sockets can do. Hao Yin et al. \cite{r14}proposed NS3-AI to connect AI algorithms and NS simulators with data interaction. NS3-AI uses shared memory to achieve an artificial combination of intelligent algorithms and network simulation. The use of shared memory dramatically reduces the time consumption in large-scale data transmission. However, there are many pitfalls in using shared memory, which can easily lead to program crashes. Meanwhile, internal process communication increases complexity of the simulator. To solve the above problems, we construct a  discrete-event-based simulator for distributed deep learning (sim4DistrDL).

\section{SYSTEM ARCHITECTURE}
The architecture of sim4DistrDL, as depicted in Fig. 2, consists of two modules, a network simulation module and a deep learning module. The former simulates the caching and transmission of data in the edge network, while the latter implements  distributed deep learning based on cached data. The main contribution of this work is the design and implementation of sim4DistrDL. Data caching and computation units are included to facilitate seamless integration of communication function and deep learning function. 
In the following subsections, we describe sim4DistrDL components in detail.
\begin{figure*}[h]
\centerline{\includegraphics[width=0.75\textwidth]{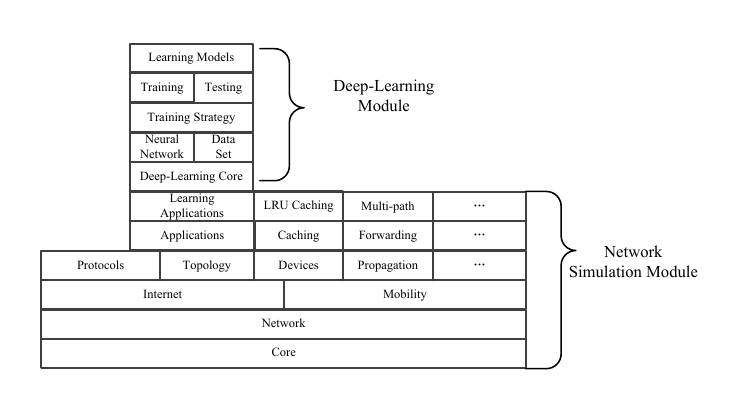}}
\vspace{-16pt}
\caption{Architecture of sim4DistrDL.}
\label{fig}
\end{figure*}

\subsection{The Overview of the Design}\label{AA}
As illustrated in Fig. 2, components only have dependencies on modules beneath them. The network simulation module consists of all components in network simulation module. The fundamental networking functionality of sim4DistrDL is supported by the core components. 
The core component includes a smart pointer, callback, tracking, logging, random variables events, scheduler, time calculation, and so on. In addition, the primary functionality of the network simulation module, \emph{i.e.}, data transmission is implemented in the network component. It implements multiple functions, node definition, network device, MAC address type (IPv4), queue, socket, IPv4 / IPv6 address, packet, etc. The upper components include Internet component and mobility component, i.e., channels to support data transmission. Above them, protocols component, topology component, device component and propagation component are provided to support network topology building and implement IP-based transmission. To provide space for the training data, we study different caching strategies (\emph{e.g.,} LRU) and embed them on device component. In addition, application component  is the container for different applications, including deep learning applications. The whole deep learning module is implemented based on application component.

The Deep-Learning Module includes six components: Data set, Neural network, Training Strategy, Training, Testing, and Learning models. 
Data set component is used for data processing. The utility of neural network component is to provide a tool suite for neural network construction. PerceptronLayer, ScalingLayer, UnscalingLayer, Boundinglayer, ProbabilisticLayer, LongShortMemoryLayer, RecurrentLayer, ConvolutionaLayer, PoolingLayer, PrincipalComponentsLayer can be added in a model. 
Training strategy can set the loss function and the optimization algorithm for the neural network. Training component supports the training process of the included neural network models. We include the training component into a custom application mentioned in network simulation module. Various DNN-based applications are implemented based on this type of applications. Testing component is another type of custom applications. In this way, learning models can be deployed and used in a simulated edge network.

\subsection{Network Simulation Module}
The network simulation module is a vital part of our framework since it is used to implement an environment for distributed learning. Data transmission and caching are supported by this module. 

\subsubsection{Core Components}
Discrete event scheduling can fulfill the needs of the modern networking research. Therefore, we design our simulator based on discrete event scheduling. Meanwhile, the core concepts  used in network simulation module can be well mapped to the natural learning environment. 
\begin{itemize}
    \item Nodes are the basic entities in simulations. Various components can be embed in nodes, e.g, applications, LRU-based caching. It's a container for applications, channels, and devices. 
    \item The application in the simulation environment runs on a node. It is necessary to abstract the simulated  program as an application. 
    \item The Channel class provides methods for managing communication objects and connections among nodes.
    \item The Device class offers multiple strategies for  channel management.
\end{itemize}

Besides the above, leveraging our simulator, users can set simulation typologies of the created nodes, simulate the TCP/IP protocol stack (physical, link, network, transport, application layer protocols), trace and collect simulation data and visualize the simulation results.

To support deep learning with the simulation framework, we designed and implemented a data caching component with different replacement strategies. In this way, the data required for model training can be cached in nodes. 

\subsubsection{Caching Component with Different Replacement Strategies}

Training data is indispensable for model training. To facilitate simulation of distributed learning, a component to support in-network caching is desired. 
Here, we take a caching design based on LRU (Least Recently Used) as an example. LRU is the most popular cache replacement algorithm. It is core idea is ``If the data has been accessed recently, its probability of being accessed in the future will also be higher". A LRU-based caching component is designed and included in the network simulation module. 

The component uses a linked list to cache data. The least recently used data is placed at the head of the linked list. In the implementation process of the caching component, the size of the cache will be set up. The addition, deletion and search functions of cached data are implemented on this list as follows:
\begin{itemize}
 \item To cache data, we first judge whether the number of the cached data has reached the maximum capacity of the cache. If it has reached the maximum capacity, delete the last data in the linked list and store the new data at the head of the linked list. Otherwise, the new received data is directly stored at the head of the linked list.
 \item In the removing process of the cached data, the last data in the linked list will be deleted.
 \item To look up a specific data, the index of the found data is returned. Then the data is placed at the head of the linked list. Otherwise, an error is returned.
\end{itemize}

This type of caching components is used to support model training, so they will be used together with the learning module (see the following section).  

\subsection{Deep Learning Module}
The deep learning module leverage a standardized deep learning lib to support model training and testing. Deep learning module helps users build various deep learning models. 
We cross compile network simulation module and deep learning module to guarantee their interacts. This is essential for the research on the application of deep learning algorithms in the edge network. All utility classes of network simulation module are exposed to the deep learning module for the distributed learning purpose.
 
These functions of deep learning module are achieved in some classes. The abstract class, LossIndex, includes the definition of loss index term and a regularization term of the loss function. The OptimizationAlgorithm class is used for selecting training algorithms for the neural network. There are some classes that inherit members of the LossIndex class to calculating errors, \emph{e.g.}, mean square error, that need to be set before model training. The neural network training algorithms (e.g., stochastic gradient descent algorithms) is derived by the OptimizationAlgorithm class. The basic classes, ModelSelection, includes two subclasses, InputSelection class and NeurosSelection class. Among them, InputSelection represents the input selection algorithms of the neural network. NeuronsSelection calles ModelSelection to select available neurons.

\section{Implementation}
sim4DistrDL provides a way to collect and exchange information between network simulation module and deep learning module, and enable the deep learning in the network simulation. Actually, sim4DistrDL is a framework that consists of two components (network simulation module and the deep learning module). We have released the source code of sim4DistrDL in an online repository$\footnote{Online repository: \href{https://github.com/Lxy85/sim4DistrDL.git}{https://github.com/Lxy85/sim4DistrDL.git}}$.
The simulator simplifies the tasks of the simulation development of distributed deep learning in networking environments. In the following subsections, we describe the implementation details, as well as the deployment of sim4DistrDL. 


\subsection{Implementation of the LRU-based Caching}

We implement LRU-based Caching on the existing core components. Use the waf to automatically compile this new module. The implementation of the LRU-based Caching module in the simulation environment includes the following four steps:

\begin{enumerate}
\item LRU-based Caching module construction

We need to create a new path to provide the placement of the caching component. Use terminal commands to customize the new directory as follows:

\begin{itemize}
\item ./utils/create-module.py src/Lru
\end{itemize}

In this command, parameter Lru is the name of the new component. The create-module.py command helps to create a directory structure. The system will automatically create a sub-directory named after Lru. Its internal structure includes sub-component and  the configuration file (wscript).
\item Configuration

After the new component is created, the wscript configuration file needs to be modified.  It is used to register the source code contained in the project and reference other components.

In the wscript file, we need to set other modules in the simulation module that the caching module depends on. In addition, we need to use module.source function and headers.source function to declare the included cc files and h files, respectively.
\item Add LRU-based Caching component to the network simulation module

After the component is established, it is added to the network simulation module to facilitate the use of cached data in the simulation scenario. To add the LRU-based caching component to the simulation environment, we need to use waf to recompile and configure the all modules in network simulation module.

\begin{itemize}
\item ./waf configure
\item ./waf build
\end{itemize}

The command contains the shared libraries and header files used for compiling files, and the path that must be entered when compiling scripts with a waf tool.
\item Deploy LRU-based caching component on nodes

After the cache module is successfully implemented, it needs to be deployed in the simulation node. Its header file is included in the application class. Besides, LRU-based Caching component needs to be added as a member variable of Node class so that in-network cache can be deployed in  the simulation node. 
\end{enumerate}

In this way, the LRU-based caching component is compiled, included in the  network simulation module. 

\subsection{Implement of the deep learning module}
Most DNN models rely on open-source frameworks (Pytorch, Tensorflow, etc.). Considering the resource consumption of these frameworks, this poses specific difficulties for building DNN models and verifying algorithms in a simulation. A deep learning lib (OpenNN\cite{r15}) is include to handle this problem. That does not rely on third-party frameworks, but used as a lib. It includes some popular DNN models, \emph{e.g.}, MLP. Other popular models can be added to the deep learning lib. 
In the following part of this section, we take AlexNet as an example to present how to import a model into the deep learning module. 
\begin{itemize}
\item Use a name space compatible to the lib.
\item Include all the header files that implemented AlexNet model into the total header file of the lib.
\end{itemize}

\subsection{Deployment of sim4DistrDL}
To enable a simulation scenario, one needs to integrate the designed deep learning module and network simulation module, and then encapsulate the model training into an application. The specific deployment process is as follows:
\begin{enumerate}
\item The Eigen library is a C++ template library for linear operations, which can support matrix and vector operations, numerical analysis, and other algorithms. The deep learning module contains a large number of matrix operations, so we need to use Eigen's template library to enhance our simulator. To import the Eigen in the deep learning module into the network simulation module, is the preliminary step of the joint compilation of the deep learning module and the network simulation module.
\item To integrate the deep learning module and the network simulation module, the code of the deep learning module needs to be public to the network simulation module. Additionally, we need to set up the configuration file (wscript) in the network simulation module to add the head file of the deep learning lib in the network simulation module, more specifically, add it into an application class. The configuration code is:

\begin{itemize}
\item def build(bld):
\item bld.stlib(“deep{\_}learning”)
\item module.uselib = ‘deep{\_}learning’
\item module.source = ‘deep{\_}learning/**/*.cpp’
\item module.full{\_}headers = ‘deep{\_}learning/**/*.cpp’
\item module.full{\_}headers = ‘deep{\_}learning/**/*.cpp’
\end{itemize}
The above code is used to add a static deep learning library to the configuration file of NS-3 (wscript), including the corresponding header and source files.


\end{enumerate}
After completing the above steps, a deep learning library has been successfully embed in the network simulation module.  
\section{Application Cases}
In this section, we present an application case of the proposed simulator. We first introduce the network topology, the construction of the neural network, the steps of the simulation, and some specific implementation details. Then, the simulation results are provided.
\subsection{Simulation Setup}

\begin{figure*}[h]
\centerline{\includegraphics[width=\textwidth]{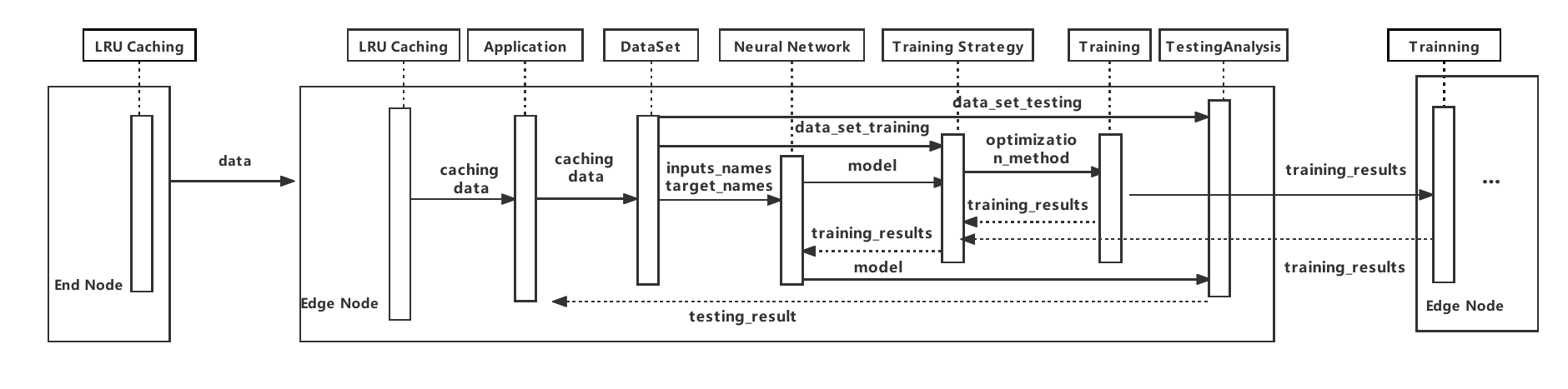}}
\vspace{-15pt}
\caption{Example of a sequence diagram.}
\label{fig3}
\end{figure*}

The topology used in the simulation process is a tree topology. It includes three connected edge nodes and their adjacent end nodes. The edge nodes and end nodes are connected through gigabit links. End devices are used for caching data, while edge nodes are used for training model. Data is transmitted among nodes through Peer-to-Peer (P2P) protocol, which is the fundamental network technology to support data transmission in Internet. 
The data required for model training are all released by the end node, and sent to the neighboring edge node. The edge computing node uses the received data and the data in its own cache to train DNN models in a collaborative way.  

We adopt a CNN model (AlexNet) and a multiple layer perceptron (MLP) as examples to present the simulation process. First of all, by using the deep learning module, these two DNN models are implemented. 
After that, the distributed training environment is configured in the simulator as the follows:

\begin{enumerate}
\item The header file of network simulation module and that of the designed cache component is included in a simulation script. In the script, we use the Create() function in the NodeContainer class to create the required 6 nodes, and then set up the links among these nodes. Initialize the cache size for each node. The cache size is set to 3000.
\item Install the protocol stacks for the created nodes, while to configure the IP address for each node.
\item Set the data requester and source. The requester are the edge nodes, and the data source nodes are the end nodes. The edge nodes use the callback function to handle the data requested from the end nodes.
\item To facilitate neural network training, we first encapsulate the model into an application. We include the header file of the deep learning module in the application. In addition, the training parameters of the model are configured, such as the maximum training round, the threshold to guarantee model converged, etc. The stochastic gradient descent algorithm is selected in the training strategy class.
\item Install the application in the edge nodes, and start the simulation. 
\end{enumerate}

To achieve a collaborative training, every edge node needs to judge whether the locally cached data is enough before training. If not, a collaborative training is started. Not only training data but also training result for each epoch will be transmitted among nodes.

\subsection{Simulation Results}

\begin{figure}[htbp]
\centerline{\includegraphics[width=0.9\linewidth]{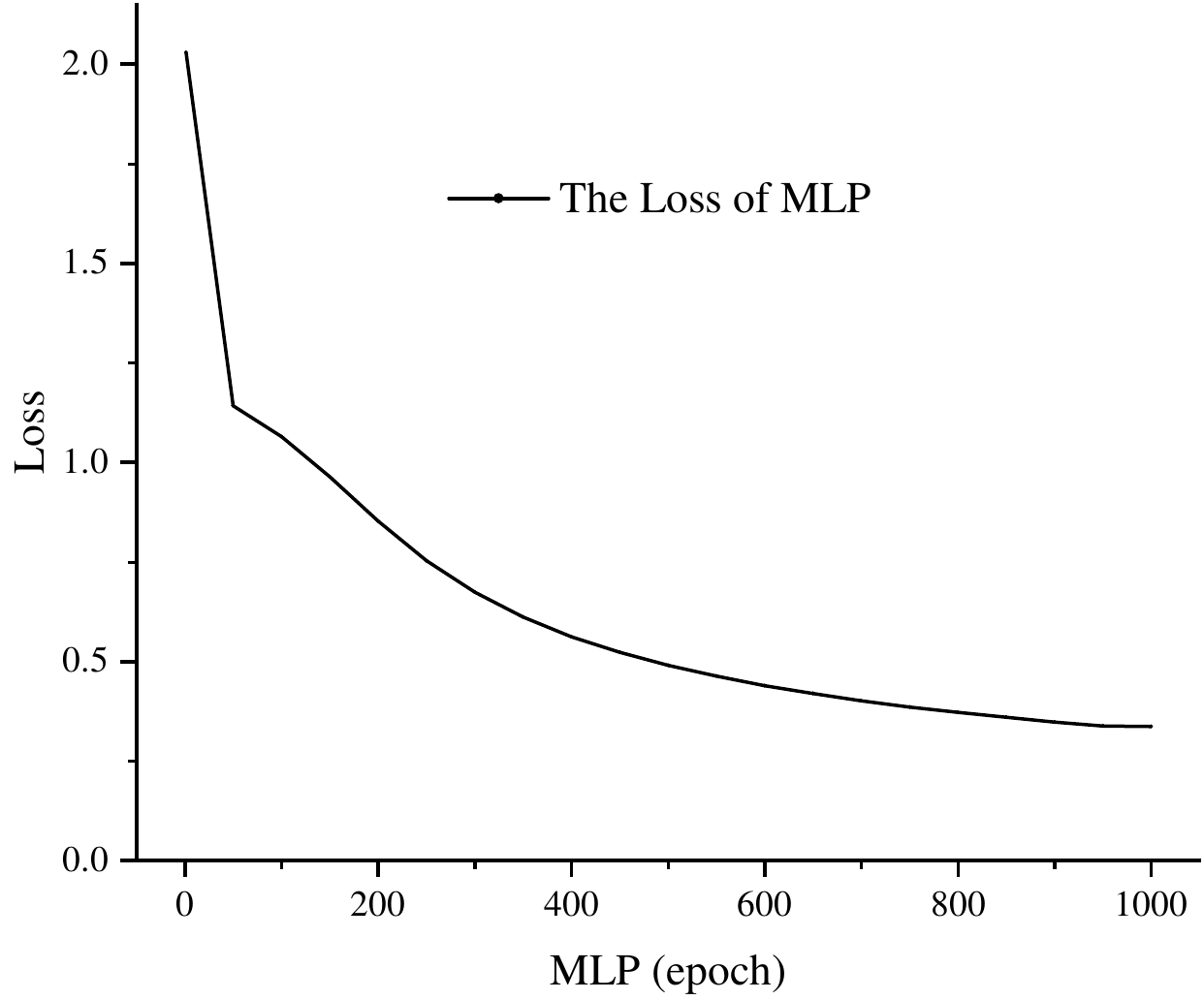}}
\vspace{-8pt}
\caption{Distributed training of multiple layer perceptron}
\label{fig4}
\end{figure}
\begin{figure}[htbp]
\centerline{\includegraphics[width=0.9\linewidth]{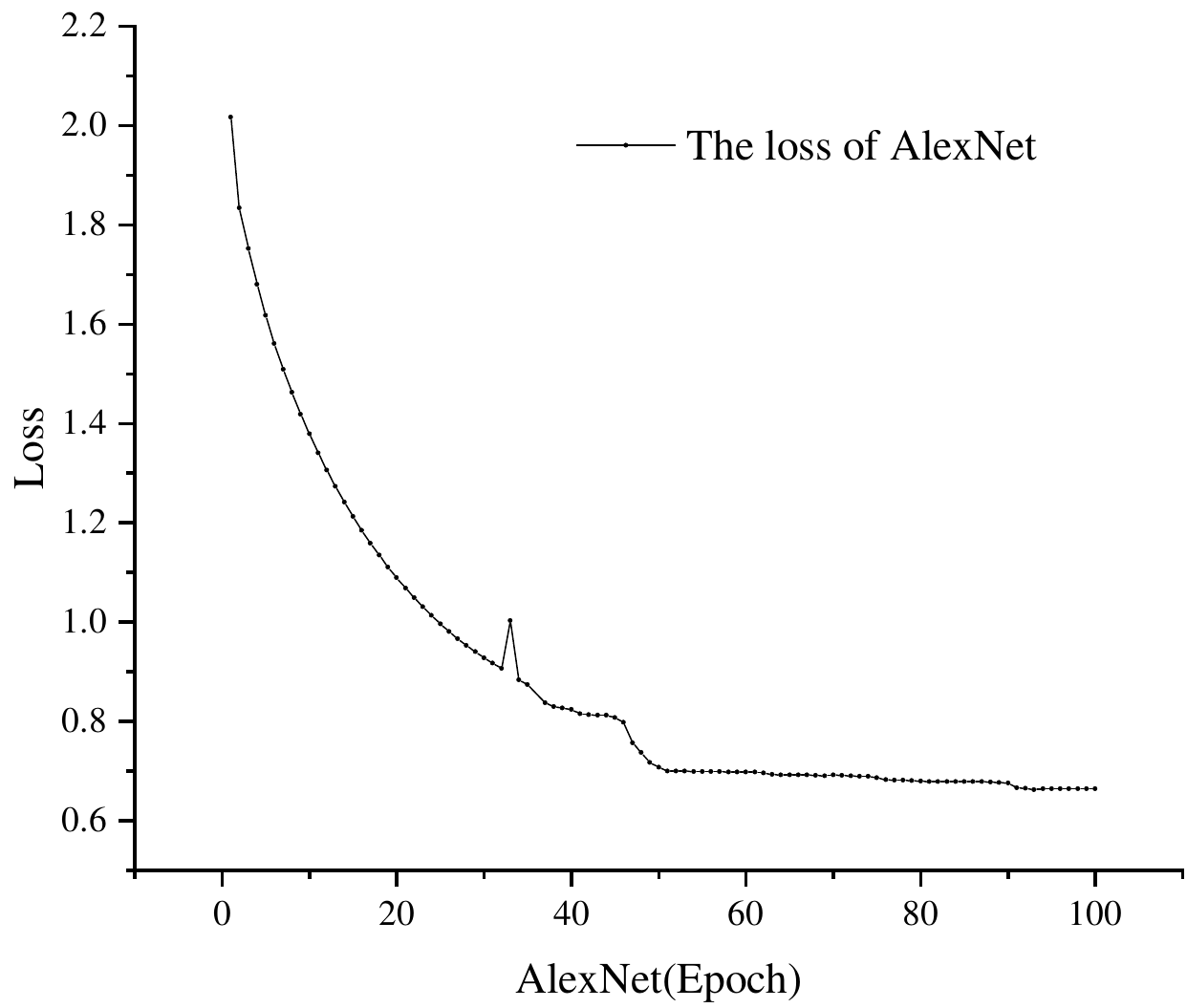}}
\vspace{-8pt}
\caption{Distributed training of CNN model (Alexnet)}
\label{fig5}
\end{figure}

A textual dataset is used to train the MLP model, and an image dataset is used by the Alexnet model.  
The values of the loss function in terms of different models (MLP and Alexnet) are depicted in Fig.4 and Fig.5, respectively. When the epoch increases, the loss values decrease. Both MLP and AlexNet can achieve convergence with different latency. These results prove that our framework is generic and can be used in identify various training processes of DNN models.

\section{CONCLUSIONS}
In this paper, we presented a discrete event simulator, sim4DistrDL, which simplifies the simulation of deep learning in a distributed environment. Its design is achieved by interconnecting the deep learning module with the network simulation module. In addition, we provide its implementation details of model implementation and deployment. As the framework is generic, the source code has been provided, and the community can use it in a variety of distributed deep learning in the edge network. Providing a prerequisite for the configuration of distributed intelligent applications, a sim4DistrDL-based simulation can give a learning environment similar to that in the real TCP/IP network, including data transmission, caching, and model training. In the near future, we plan to include more support to types of neural network models.

\bibliographystyle{unsrt}
\end{document}